\documentclass[runningheads]{llncs}
\usepackage[T1]{fontenc}
\usepackage{amsmath}
\usepackage{graphicx, caption, subcaption}
\usepackage{multirow}
\usepackage[table]{xcolor}
\definecolor{light-gray}{gray}{0.95}
\definecolor{light-blue}{rgb}{0.71,0.87,0.91}

\begin{document}

\title{Fully Automated CTC Detection, Segmentation and Classification for Multi-Channel IF Imaging}
\titlerunning{Automated CTC Detection, Segmentation and Classification}
\author{Evan Schwab, Bharat Annaldas, Nisha Ramesh, Anna Lundberg, Vishal Shelke, Xinran Xu, Cole Gilbertson, Jiyun Byun, Ernest T. Lam}
%index{Schwab, Evan}
%index{Annaldas, Bharat}
%index{Ramesh, Nisha}
%index{Lundberg, Anna}
%index{Shelke, Vishal}
%index{Xu, Xinran}
%index{Gilbertson, Cole}
%index{Byun, Jiyun}
%index{Lam, Ernest}

%\author{Anonymous}
%
\authorrunning{E. Schwab et al.}
%\authorrunning{Anonymous}
\institute{Epic Sciences, San Diego, CA, USA}
%\institute{Anonymous}
%
\maketitle              % typeset the header of the contribution
\begin{abstract}
Liquid biopsies (eg., blood draws) offer a less invasive and non-localized alternative to tissue biopsies for monitoring the progression of metastatic breast cancer (mBCa). Immunofluoresence (IF) microscopy is a tool to image and analyze millions of blood cells in a patient sample. By detecting and genetically sequencing circulating tumor cells (CTCs) in the blood, personalized treatment plans are achievable for various cancer subtypes. However, CTCs are rare (about 1 in 2M), making manual CTC detection very difficult.  In addition, clinicians rely on quantitative cellular biomarkers to manually classify CTCs. This requires prior tasks of cell detection, segmentation and feature extraction. To assist clinicians, we have developed a fully automated machine learning-based production-level pipeline to efficiently detect, segment and classify CTCs in multi-channel IF images. We achieve over $99\%$ sensitivity and $97\%$ specificity on 9,533 cells from 15 mBCa patients. Our pipeline has been successfully deployed on real mBCa patients, reducing a patient average of 14M detected cells to only 335 CTC candidates for manual review.

%Digital features like channel specific cell intensity are necessary for experts to manually classify a cell as a CTC and requires automated cell detection and segmentation.

%The abstract should briefly summarize the contents of the paper in 15--250 words.

\end{abstract}
\section{Introduction}
%DefineMBC is a patient-specific test for Metastatic Breast Cancer (MBCa), derived from a liquid biopsy (ie, blood draw) \cite{}. 
Liquid biopsies (eg., blood draws) offer a less invasive and non-localized alternative to tissue biopsies for a more continuous monitoring of metastatic breast cancer (mBCa). Immunofluoresence (IF) microscopy is a tool to image and analyze millions of cells in a sample of blood. Circulating tumor cells (CTCs) in the blood are indicative of metastasis and the combination of protein biomarker and single-cell genetic analyses of CTCs provides clinicians with a comprehensive cancer profile for mBCa patients from a single blood draw \cite{vidlarova2023recent,werner2015analytical}.

In mBCa patients, CTCs average 1 in every 2M cells \cite{werner2015analytical}, %\footnote{This is based on a verification study of 81 mBCa patient samples.})
motivating the need for machine learning (ML) to reduce the burden of manual classification by automatically detecting and classifying CTCs. Deep learning has been developed for classifying CTCs in multiple IF modalities  \cite{guo2022circulating,hashimoto2023automatic,park2024classification,shen2023automatic,tsuji2020detection,zeune2020deep,zhang2019automated}. However, these methods fall short of providing a fully automated pipeline to detect and classify CTCs to present for manual review. Furthermore, deep learning loses the utility and interpretability of biomarker features, which clinicians already rely on for manual classification.

\begin{figure}[ht!]
    \centering
    \includegraphics[width=1\textwidth]{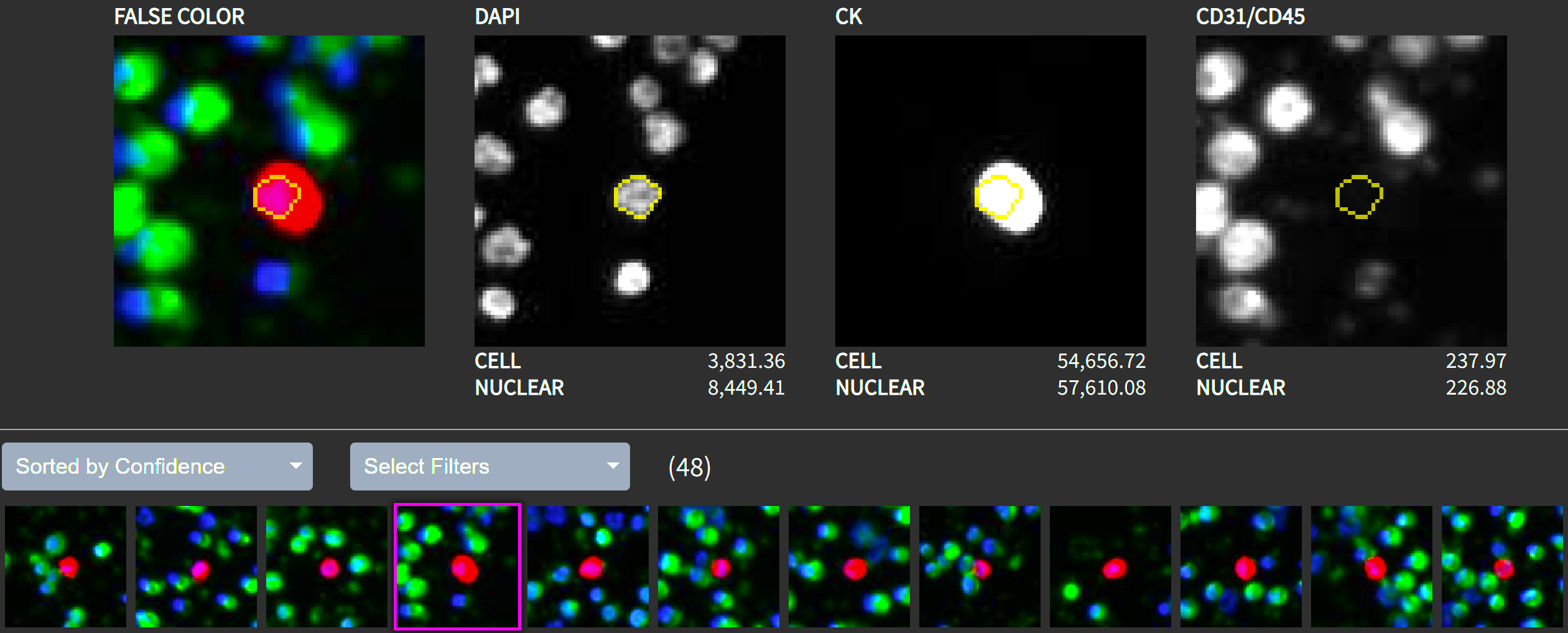}
    \caption{Example user interface of ML classified CTC candidates output by our pipeline and presented to clinicians (bottom row thumbnails). One thumbnail is selected for review and cell and nuclear MFI values reported per channel. The yellow nuclear mask is overlayed on each channel and the color composite. %In this patient slide, our pipeline detected 48 CTC candidates out of a total of 3M cells.
    }
    \label{fig:viewer}
\end{figure}
In this work, we have developed a fully automated ML-based production-level pipeline to efficiently detect, segment, and classify mBCa CTCs in multi-channel IF images to present to clinicians for final confirmation and genetic sequencing. Our automated pipeline, which we call \textbf{BR}east cancer \textbf{I}maging \textbf{A}lgorithm (BRIA), combines image processing, deep learning and interpretable feature-based ML. BRIA has been fully deployed at Epic Sciences within a comprehensive clinical workflow that integrates patient sample collection, slide preparation, fluorescence scanning, cloud-based analysis using the proposed pipeline, database management, quality control, and reporting through a proprietary clinical viewer that interacts with data sources through an API.

In Sec.~\ref{sec:background} we provide a background on IF imaging and the clinical workflow. Then, in Sec.~\ref{sec:methods} we present each component of the BRIA pipeline and share our ground truth labeling and experimental results in Sec.~\ref{sec:experiments}.

\section{Background}
\label{sec:background}

\begin{figure}
    \centering
    \includegraphics[width=1\textwidth]{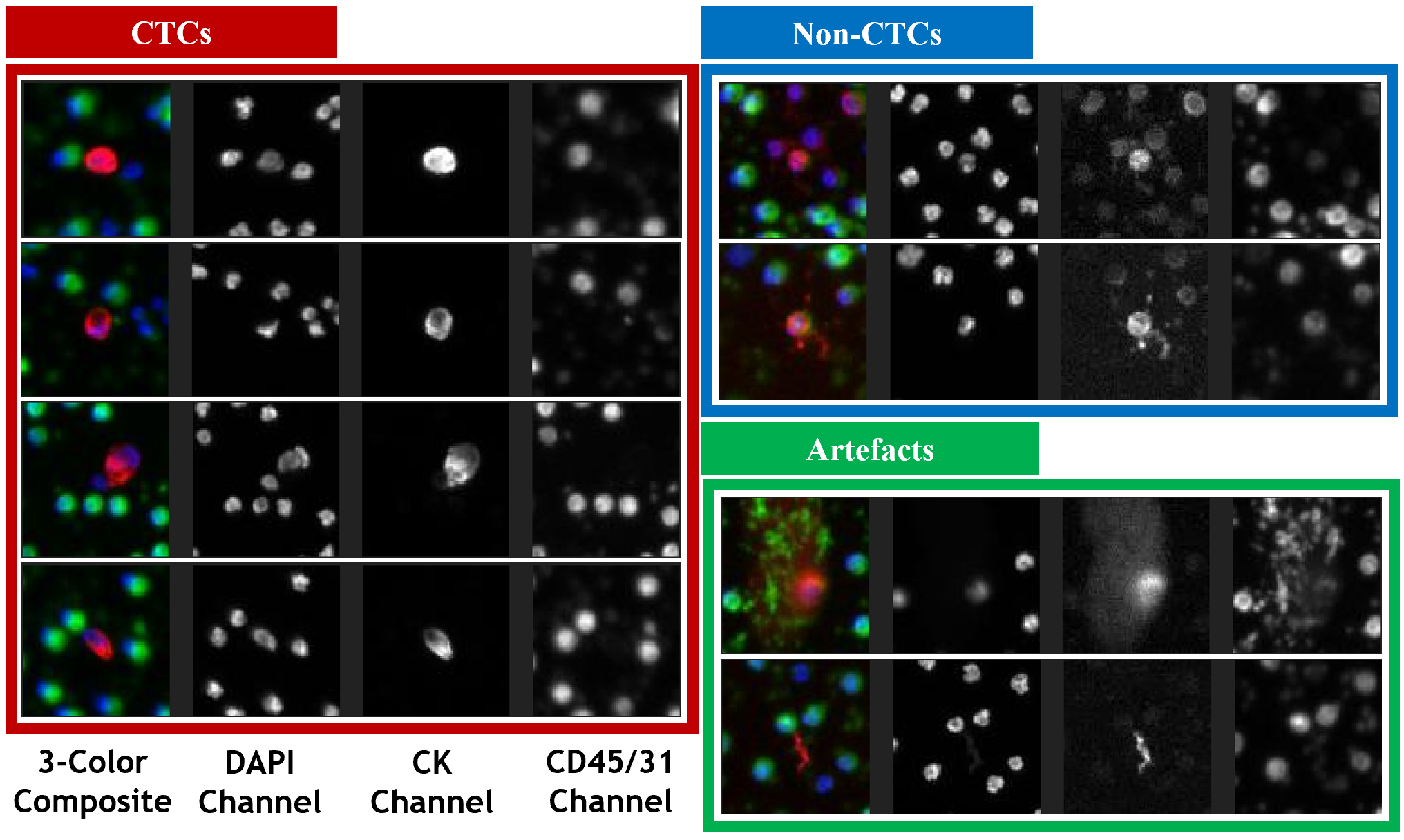}
        \caption{Example CTCs, non-CTCs, and artefacts in 3-channel IF images with a color composite. CTCs have prominent CK signal and low CD45/31. Some Non-CTCs are visually similar which makes manual classification difficult.}
    \label{fig:ctcs}
\end{figure}

Widefield fluorescence microscopy is an IF imaging technique used to capture 2D images of blood samples collected on whole slides \cite{scher2021development,wilson_2017}.
%\cite{anonymous,wilson_2017}. 
Fluorescent dyes are applied in an assay system to capture specific biomarkers of interest and are digitized in multiple imaging channels. At Epic Sciences, we have developed a clinical diagnostic test for mBCa, called DefineMBC, which uses IF imaging as one of the core components for protein biomarker expression analysis. The DefineMBC assay includes DAPI, CK, and CD45/31 channels which are designed to visualize cell components and provide inclusionary and exclusionary biomarkers for CTCs.\footnote{BRIA is agnostic to additional BCa biomarker channels like HER2 or ER.} The DAPI stain (4',6-diamidino-2-phenylindole) is used to highlight the nucleus of the cell. The CK stain (Cytokeratin) is an indicator of CTCs and CD45/31\footnote{Cluster of differentiation (CD) 31 is a protein associated with white blood cells and endothelial cells. CD 45 is a protein marker of leukocyte lineage.} is an indicator of non-CTCs like white blood cells. Therefore, high CK and low CD45/31 values are indicative of CTCs (see Fig.~\ref{fig:ctcs} for examples). 

These channel values are summarized using statistical features such as the mean florescence intensity (MFI), ie. the average pixel intensity within an object like the cell or nucleus, for each channel (e.g., Nuclear CK MFI). To compute these features, each individual cell on a slide of roughly 3M cells must be detected and cropped as individual thumbnail images.  Then the cell and nucleus need to be segmented to compute MFIs for each channel. %footnote{To avoid segmentation of hundreds of millions of patient cells, a low CK$+$ threshold is first applied to each thumbnail to filter most obvious non-CTCs and artefacts.} Then nuclear and cell segmentation is applied and MFIs computed. 
In the absence of ML, the standard practice to classify CTCs is to first apply biomarker thresholds (eg. Nuclear CK MFI $>\!269$ and Nuclear CD45/31 MFI $\leq\!3000$, etc.) to filter 99\% of non-CTCs and artefacts in our data. However, validating biomarker thresholds is a challenging task and clinicians are still left with 10's of thousands of CTC candidates per patient to sort through.

%To arrive at a manual CTC classification, multiple image processing steps are necessary to extract MFIs. First, each individual cell on a slide of roughly 3 million cells must be detected and cropped as individual thumbnail images. %\footnote{Whole-slide IF images are too large to process fully so they are first divided into a grid of smaller field of views (FOVs) for processing and quality control.} 
%Then the nucleus needs to be segmented from the DAPI channel and the CK and CD45/CD31 MFIs calculated from the segmented mask. However, MFIs alone are not sufficient to classify CTCs from non-CTCs like WBCs, endothelial cells, or non-cellular artefacts like flares. 

Within this workflow, clinicians manually classify CTCs based on MFI values and the visual inspection of features like cell size, shape, texture and other biomarkers. Because clinicians already rely heuristically on a list of qualitative cellular features, we are motivated to quantify these biomarkers and utilize interpretable feature-based ML instead of deep learning to classify CTCs. %We extract a number of morphology, intensity and texture features from the cell, nucleus, and thumbnail images as important step of our pipeline.

The BRIA pipeline assists clinicians by automating these steps of cell detection, nuclear and cell segmentation, feature extraction and CTC classification to present CTC candidates to clinicians for final review. Figure~\ref{fig:viewer} shows a user interface of CTC candidates presented to clinicians for manual review. For this slide example, the pipeline reduced 3M total cells to 48 CTC candidates instead of what would be 10s of thousands of candidates to sort through.

%we present novel applications of these techniques to the uniquely challenging case of widefield multi-channel IF imaging of CTCs. 

%deep learning
%mao2016deep, tsuji2020detection, zeune2020deep, guo2022circulating, hashimoto2023automatic, shen2023automatic, park2024classification

%rcnn
%zhang2019automated

%zeune2017quantifying
%de2018classification

%wang2020label

%review
%vidlarova2023recent, 

%We describe the image processing steps of the \textcolor{red}{BRIA} pipeline in the next section.

%A number of quality control metrics are applied to determine if a slide is ready for processing or should be re-scanned.
%In this application, we strive for high sensitivity, ie it's important to not miss any CTCs. While we can be more liberal on specificity, too many false positives will create more burden for the human reviewers.

\section{Methods}
\label{sec:methods}
\begin{figure}[ht!]
    \centering
    \includegraphics[width=1\textwidth]{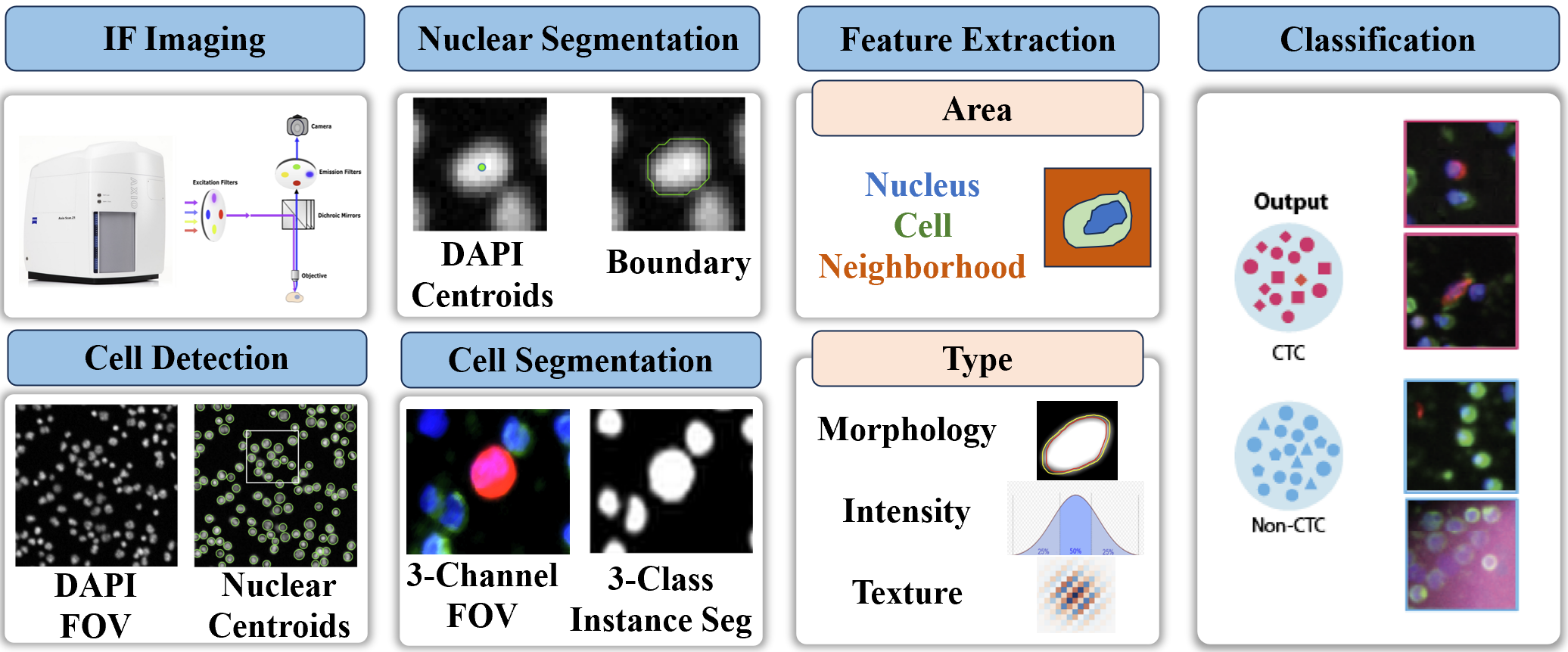}
    \caption{BRIA Pipeline Overview. After IF image acquisition and slide processing, the main algorithmic steps include Cell Detection, Nuclear and Cell Segmentation, Feature Extraction, and CTC Classification.}
    \label{fig:bria}
\end{figure}

We outline the main steps of the BRIA pipeline leading up to CTC classification in Sec~\ref{sec:detection} cell detection, \ref{sec:nucleus_segmentation} nucleus segmentation, \ref{sec:cell_segmentation} cell segmentation, and \ref{sec:feature_extraction} feature extraction. These steps are also required for manual classification in the absence of an ML CTC classifier. See Figure~\ref{fig:bria} for a visual overview. %FOVs have $2\%$ overlap to avoid cutting into individual cells. %( GIVE Pixel dim, File size), so they are first divided into a grid of smaller field of views (FOVs) for which our algorithms are applied in parallel. 

\subsection{Cell Detection}
\label{sec:detection}
The first step of BRIA is to efficiently detect each individual cell center and crop thumbnail images for each cell. Whole slides are too large to image fully and so a grid of 588 ($14\!\times\!42$) fields of view (FOVs) of size $2040\!\times\!2040$ pixels are imaged and processed in parallel and stitched together.
%The DAPI channel which highlights the cell nuclei is used for cell detection.
%In order to locate cells for classification, the first step of the pipeline is to automatically locate the coordinates of individual cell centers and produce a thumbnail region of interest (ROI) around each cell. Cell detection is designed to locate the centroids of nuclei, referred to as seeds. This step is motivated by quickly locating cells from a large-scale dataset to initialize the more sophisticated segmentation steps and to produce a set of thumbnail region of interest (ROI) images centered at each detected seed. 
We compared three classical cell detection methods including watershed, radial symmetry \cite{schmitt08} and the Laplacian of Gaussian (LoG) \cite{Stegmaier14}. These are applied to the DAPI channel which highlights the nucleus of each cell (CTCs and non-CTCs alike).

%We experimented with various LoG parameters including $\sigma_{min}$ and $\sigma_{max}$ that determine how well the filter approximates the minimum and maximum size of nuclei, and $n$ which determines how many intermediate sizes of cells to detect. 

For validation, the centroid coordinates of 24,110 cells from 5 FOVs were manually identified. We measured performance based on the cell count output by the detection algorithm as well as the average distance between the ground truth and estimated centroids. With an exhaustive parameter search, the best performing algorithm was the LoG with cell count F1 score of 0.997 and a 1.12 $\mu$m average distance between estimated and ground truth centroids. This is an acceptable error compared to the average radius of a cell nucleus of about 5 $\mu$m and whole cell size of up to 15 $\mu$m. Small $24\!\times\!24$ pixel thumbnail images are cropped around the centroids based on the average size of nuclei. In total, our cell detection algorithm takes 10 min to detect $\sim3$M cells from a single slide. % and cell size between 3.5 $\mu$m to 15 $\mu$m \cite{wbc-wiki}. % (CTCs are generally larger than WBCs). 

%The summary of F-measure is shown in Figure \ref{fig:f-measure-benchmark}. All three (3) methods performed pretty well. The multi-scale LoG and radial symmetry based methods both provide consistent results regardless of local/global intensity variations. As expected, watershed segmentation is sensitive to intensity variations, thus it is more susceptible to imaging noise such as out of focus or lint/dust on the slide. Multi-scale LoG-based segmentation performs two (2) times faster than the radial symmetry based method yet achieves better accuracy 

\subsection{Nuclear Segmentation}
\label{sec:nucleus_segmentation}
%\label{sec:segmentation}
Once a cell is detected, segmentation of the nucleus is important for extracting features like nuclear MFI for each channel. 
%Based on our particular multi-channel IF data, we apply different techniques for nucleus and cell segmentation. 
Since the DAPI channel is already designed to highlight nuclei, nuclear segmentation can be accomplished with classical image processing. 
%Alternatively, signals for the whole cell are represented differently across channels. Therefore, we invoke a separate multi-channel approach for whole cell segmentation.
%\subsubsection{Nucleus Segmentation.}
%\label{sec:nucleus_segmentation}
%For nucleus segmentation, 
%We apply a gradient-based image processing approach to the nuclei within the DAPI channel. 
Given each thumbnail image, we compute the normal vector field of each pixel to the center and the gradient vector field. These two fields are multiplied using a Gaussian weighted dot product with a radius parameter output by the detection algorithm. This results in a transformed image that isolates the cell of interest. We compared the watershed method and Otsu's method \cite{otsu79} applied the transformed image to segment the nucleus. 

To evaluate performance, the nuclei of the same set of 24,110 cells that were used for the cell detection ground truth were also manually segmented by 40 expert annotators. On a sample of 25 nuclei, we used the Simultaneous Truth and Performance Level Estimation (STAPLE) \cite{staple} to measure the concordance between annotators and found 93\% sensitivity for segmenting the nucleus correctly. Watershed achieved F1=0.83 while Otsu's method achieved F1=0.934 which is a good performance in comparison to human annotators. With parallelization the nuclear segmentation step takes an average of 10 min per slide.

\subsection{Cell Segmentation}

%\subsubsection{Cell Segmentation.}
\label{sec:cell_segmentation}
For whole cell segmentation we utilize all three IF channels to account for signal variation  across each channel for both CTCs and non-CTCs alike. To address the scarcity of CTCs in our data, simulated montages of CTCs are created to increase their occurrence by placing manually segmented CTCs at randomly generated coordinates on a simulated background image based on real IF images.  

We utilize a 3-channel U-Net \cite{unet} to segment cells within overlapping $512\!\times\!512$ image patches within each FOV. Overlap was enforced to avoid cutting cells at the edge of a patch. To accelerate this process we use our cell detection output to discard patches that do not contain any cells. The outputs of the U-Net are pixel-level probability maps for three classes: cell, boundary, and background. Including the boundary as a third class proved useful in separating cell clusters. The probability maps for each overlapping patch are merged into a single FOV by taking the maximum among the cell, boundary, and inverted background probability maps. Then instance segmentation is performed using watershed to identify the masks of each individual cell. These masks are mapped back to each thumbnail using the cell centroids.

Training was performed on 22 patient slides and 9 CTC montages with a total of $10,443$ manually segmented cells. The STAPLE algorithm was again used to evaluate human level segmentation performance. We gave 10 annotators 1,206 cells (including 165 CTCs) to manually segment and their concordance level was an F1=0.936. In comparison our U-Net achieved a comparable F1=0.933 averaged over 5-fold cross validation of our training set. The best performing parameters were learning rate=0.0001, dropout=0.3, weighted classes, network depth=4, epochs=300, and Adam optimizer. Our algorithm took an average of 126s per FOV and with parallelization, segmentation of an entire slide (588 FOVs, 3M cells) was accomplished in 39 min.

\subsection{Feature Extraction}
\label{sec:feature_extraction}

Feature-based ML is important for clinicians to better interpret classification results. Since clinicians already rely on quantitative values like nuclear MFI and visually inspect cell morphology and texture, we are motivated to extract additional morphology, intensity, and texture features from the nucleus, whole cell, and entire thumbnail image. In total we extract 122 features which are summarized with equations in Supp. Table 1.
 %Intensity and morphology features are calculated using nuclear and cell segmentation masks, while texture features use a mixture of region of interest and masks to compute features.

%\subsubsection{Morphology.}

We extract 8 morphology features, 4 each from the nuclear and cell masks: size, roundness, elongation and the first Hu moment \cite{moments_hu}, which captures more subtle shape variability. (BCa CTCs are often larger than non-CTCs.) 
We next compute a total of 44 intensity features from nuclear and cell masks across the three channels: MFI, lower, median, and upper quartiles, interquartile range, as well as Pearson's correlations and Ranked-Weighted Co-Localizations \cite{Singan2011} between channels. We also extract CK specific features like CK+ ratio, defined as the number of CK pixels that are greater than a cutoff value, divided by the area of the mask. CTCs will exhibit a higher CK+ ratio. We also compute the mean and standard deviation of pixels in the entire CK channel thumbnail which may help eliminate CK+ artefacts like flares.

Finally, we extract 70 texture features. First, a 32D Gabor feature vector is constructed using the mean and standard deviation of filtered images for 16 parameter combinations: $\theta\!=\!0^{\circ},45^{\circ},90^{\circ},135^{\circ}, \lambda\!=\!0.1, 0.4$, and $\sigma\!=\!1, 3$, selected based on cell size. Gabor features identify frequency changes in an image at various orientations and sizes. This helps to identify small dye-aggregates, flares as well as CTCs in an image. Next, a 32D Laws \cite{laws} feature vector is constructed by ordered multiplications of 1D filters L5 (Level), E5 (Edge), S5 (Spot), and R5 (Ripple) to detect spatial patterns.
% \begin{figure}[h!]
%     \centering
%     \includegraphics[width=1\textwidth]{LBP_CTC_NonCTC_Artefact.PNG}
%     \caption{Local Binary Pattern (LBP) images of DAPI, CK, CD45CD31 channels for CTC, Non-CTC, and Artefact examples. LBPs reveal geometric textures. Our features comprise of the correlation and mutual information of each channel pair for each sample.}
%     \label{fig:LBP}
% \end{figure}
Finally, Local Binary Pattern (LBP) \cite{lbp} encodes edges, corners, raised areas, and lines. 
%For each pixel in an image, surrounding pixels receive a 1 if they are greater than the center pixel or 0 otherwise. The center pixel is replaced by the resulting binary value vector. 
%Figure~\ref{fig:LBP} shows example LBP channel images for a CTC, Non-CTC and artefact with varying textures. 
LBP results in a transformed image per channel and we calculate the correlation and normalized mutual information between each channel pair, resulting in 6 final texture features.

%two texture features, correlation and co-localization of pixel intensities, within the nuclear and cell masks across channels. We choose an additional seven families of features to capture texture across the ROI: Gabor filters, convolutional features, Laws energy \cite{laws}, cross channel local binary pattern, blur score, histogram of gradients \cite{hog}, and statistical mean and variance. The full description and dimension of each family of texture features is summarized in Appendix Table~\ref{table:texture_features}.

%In total, we collected 125 morphology, intensity and texture features. We next used recursive feature elimination (RFE) to find the most relevant features for CTC classification. We performed CTC classification using a linear SVM and recursively removed the features with the least importance weighting. Cross-validation was used to score different feature subsets and select the best scoring collection of features. This process resulted in a subset of 46 final features listed in Appendix Table~\ref{table:final_features}. ---------Talk about data used ----

%\subsection{CTC Classification}
%\label{sec:classification}
%The goal of the CTC classifier is to classify CTCs from Non-CTCs and artefacts. In this application, we want to optimzie for sensitivity, ie it's important to not miss any CTCs. We can be more liberal on specificity, ie allowing false positive CTCs, but too many false positives will create more burden for the human reviewers.

\section{Experiments}
\label{sec:experiments}
Once cells are detected, segmented and features are extracted using the steps of the pipeline described above, we are ready to collect and label ground truth data to train and evaluate a CTC classifier. The CTC classifier is used to replace the standard rule-based biomarker thresholding and substantially reduce the number of CTC candidates presented to clinicians for final review.

\subsection{Ground Truth Data}
\label{sec:data}
Ground truth data is collected from 15 mBCa patients using the same cell detection and segmentation steps of the pipeline.  From 15 patients, a total of 241,644,731 cells were identified by the cell detection algorithm. Then 99.8\% of non-CTCs and artefacts were filtered using the rule-based thresholds described in Section~\ref{sec:background} leaving 500,255 CTC candidates presented for manual classification. (Without a CTC classifier, this is the number of candidates clinicians would have to manually review in their workflow.)  Ground truth CTCs were labeled by four annotators and subjected to a round of adjudication for consensus. They each labeled a comparable number of non-CTCs and non-cellular artefacts. 
\begin{table}[ht]
\centering
\begin{tabular}{ | c || c | c | c | c | c | c| }
 %\multicolumn{6}{c}{Results Summary} \\
 \hline
 \rowcolor{light-blue}
 \textbf{Dataset} & \textbf{Patients} & \textbf{Labeled Samples} & \textbf{CTCs} & \textbf{Non-CTCs} & \textbf{Artefacts}\\
 \hline
 \textbf{Training}   & 7  & 4,680 & 1,667 & 1,632 & 1,381 \\
 \hline
 \textbf{Verification} & 5  & 1,931 & 340 & 882 & 709 \\
 \hline
 \textbf{Validation} & 3  & 2,922 & 1,221 & 969 & 732 \\
 \hline
 \textbf{Total} & 15  & 9,533 & 3,228 & 3,483 & 2,822 \\
 \hline
\end{tabular}
\caption{Summary of class sample counts for each dataset split.}
\label{table:data_splits}
\end{table}
The result was 9,533 labeled samples with 3,228 CTCs, 3,483 non-CTCs, and 2,822 artefacts. %Non-CTCs and artefacts will be combined into a single class, but labeling them distinctly is useful to gather diverse examples of each type. 
The ground truth data was split at the patient level into training, verification, and validation sets, stratified by number of CTCs (See Table~\ref{table:data_splits}).
%The training dataset consists of 7 patients with roughly 52\% of CTCs, resulting in 1,667 CTCs, 1,632 non-CTCs, and 1,381 artefacts. The verification dataset consists of 5 patients with roughly 11\% of the CTCs, resulting in 340 CTCs, 882 non-CTCs, and 709 artefacts.
%For validation, data from the remaining 3 patients was selected such that the number of CTCs is approximately 38\% of the total CTCs collected, resulting in 1,221 CTCs, 969 non-CTCs and 732 artefacts. 
%\usepackage{tabulary}

\subsection{CTC Classifier Training}
\label{sec:classifier}

For CTC classification, we combined the non-CTC and artefacts into a single negative class for binary classification. We evaluated four SVM models, including a linear SVM and three non-linear kernel SVMs: radial basis function (RBF), sigmoid, and polynomial. To optimize the classifier performance, the hyper-parameter tuning of $C$, $\gamma$, and polynomial degree were performed using grid search over 5-fold cross validation. All image features were normalized using min-max normalization fit on the training dataset and applied to the verification and validation sets. For our application, it is imperative for patient outcomes to miss as few CTCs as possible (false negatives). Alternatively, allowing false positive CTC candidates only increases the burden of the manual reviewers. Therefore, optimal performance hinges on maximizing sensitivity while controlling for specificity. %We therefore select a final classifier probability threshold that leads to 100\% sensitivity in the training set and apply this to our verification set. %We define our success criteria on the verification and validation sets as Sensitivity $\geq\!95\%$, Specificity $\geq\!85\%$, and overall Accuracy $\geq\!90\%$.

\begin{figure}[ht!]
    \centering
    \includegraphics[width=1\textwidth]{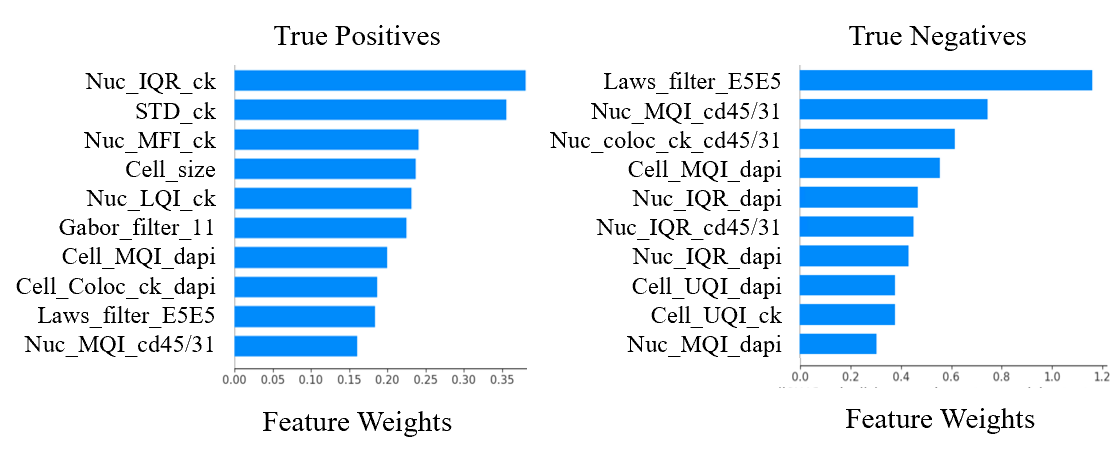}
    \caption{Top ten features weights for True Positive (CTC) and True Negative (non-CTC/artefact) classes using SHapley Additive exPlanations (SHAP) \cite{SHAP} averaged over 100 random samples per class in the verification set. (Note: Nuc\_IQR\_ck is the inter-quartile range (IQR) of pixels in the nucleus (nuc) in CK; STD\_ck is the standard deviation of CK; Cell\_Coloc\_ck\_dapi is the co-localization of pixels between CK and DAPI within the cell. See Supp. Table 1 for feature definitions.)
    }
    \label{fig:feature_weights}
\end{figure}

\begin{table}[ht!]
\centering
\begin{tabular}{ %|p{2cm}||p{1cm}|p{1cm}|p{1cm}|p{1cm}|p{1cm}|p{1cm}|p{1cm}| }
|c||c|c|c|c|c|c|c|}
 %\multicolumn{8}{c}{Results Summary} \\
 \hline
 \rowcolor{light-blue}
 \textbf{Dataset}& \textbf{True Pos.} & \textbf{True Neg.} & \textbf{False Pos.} & \textbf{False Neg.} & \textbf{Sens.} & \textbf{Spec.} & \textbf{Acc.} \\
 \hline
 \textbf{Training}   & 1,667 & 3,007 & 6 & 0 & 100\% & 99.8\% & 99.9\% \\
 \hline
 \textbf{Verification} & 340 & 1,571 & 20 & 0 & 100\% & 98.7\% & 99.0\% \\
 \hline
 \textbf{Validation} &  1,210 & 1,648 & 53 & 11 & 99.1\% & 96.9\% & 97.8\% \\
 \hline
\end{tabular}
\caption{Summary of True Positive, True Negative, False Positive, False Negative counts and Sensitivity, Specificity, and Accuracy for each dataset split.}
\label{table:results}
\end{table}

\subsection{Results}
\label{sec:results}

The SVM with RBF Kernel and hyper-parameters of $C\!=\!10$ and $\gamma\!=\!1$ scored the highest average accuracy across the 5 folds. The chosen model was retrained on the complete training set and a 0.3 probability threshold was selected to maintain 100\% sensitivity on the training and verification sets with greater than 98\% specificity. Of the 26 total false positives in the training and verification sets, 8 were actually Non-CTCs and 18 were determined to be artefacts.  Since artefacts are typically easier to inspect by humans this majority of false positives further alleviates the manual review of CTC candidates. We then applied our model to the hold-out validation set and achieved over 99\% sensitivity and 97\% specificity. See Table~\ref{table:results} for these results.

%\textcolor{red}{\textbf{Talk about False cases here. Sentence on False Negatives in Validation set and edge cases of apoptotic cells.}}

%achieving 100\% Sensitivity, 99.80\% Specificity, and 99.87\% Accuracy. 
%Of the 26 total false positives in the training and verification sets, 8 were actually Non-CTCs and 18 were determined to be artefacts. 
%We then applied the model and threshold to the verification set and achieved our success criteria with 100\% Sensitivity, 98.74\% Specificity, and 98.96\% Accuracy. 
%In total, there were 20 false positive CTCs of which 5 were actually Non-CTCs and 15 were artefacts. 
%With these results, we then applied this model to validation set achieving 99.10\% Sensitivity, 96.88\% Specificity, and 97.81\% Accuracy.
The ranked feature weights in Fig.~\ref{fig:feature_weights} indicate that CK intensity features are most important for the CTC class while DAPI and CD45/31 features are more useful for classifying non-CTCs/artefacts. This result coincides with clinical importance of CK in classifying CTCs and can be used for additional interpretation by clinicians.

Finally, we applied BRIA on the full patient data of the combined training and verification sets. With a total of 12 patients, 171M cells were detected (avg. 14M per patient). By biomarker thresholding, 401,608 CTC candidates would have been presented for manual review. In contrast, our ML CTC classifier achieved a $100\times$ reduction for a total of 4,019 CTC candidates (avg. 335 per patient) of which 2,007 are known to be true positive CTCs. Because we achieved 100\% sensitivity in the combined training and verification sets, we know that no CTCs were missed on the full patient data. This showcases the true clinical value of our ML-based pipeline to substantially reduce manual workloads. 

%\textcolor{red}{\textbf{Put these numbers is a table or figure to show comparison better. Can calculate specificity on entire dataset in both scenarios.}}

%\textcolor{blue}{In total, the complete BRIA pipeline uses XXX storage in AWS, YYY memory, and takes an average of ZZZ hours to run.}

The BRIA pipeline is configured to leverage AWS Batch jobs that run in parallel, enabling the simultaneous analysis of multiple slides. Each slide takes an average of 90 minutes to complete the analysis. The pipeline requires 32 GB of memory and up to 12 GB of temporary storage for decompressed TIFF image files, while the analysis output files, including slide QC, identified CTC MFI values in JSON format, and cell thumbnails as PNG images, require less than 50 MB of S3 storage.

\section{Conclusion}
\label{sec:conclusion}
In conclusion, we have demonstrated the clinical utility of our fully automated BRIA pipeline for detection, segmentation, feature extraction, and classification of mBCa CTCs in multi-channel IF imaging. This work is important to deliver patient-specific profiles for mBCa by assisting clinicians in detecting the rare occurrences of CTCs in liquid biopsies. Validated on a hold-out set with over 99\% sensitivity our CTC classifier detects nearly all CTCs while reducing substantial burden by presenting 100x fewer candidates for manual confirmation. With BRIA in production, future efforts will focus on continuous learning to monitor and maintain performance of CTC classification over time and expansion to additional cancer types for additional generalization and validation.

\begin{credits}
\subsubsection{\ackname} We thank our ground truth labeling teams, our medical director and the Assay Development, MTT and Clinical groups at Epic Sciences for their support in this work.

\subsubsection{\discintname}
The authors are current or past employees of Epic Sciences and may own company stock options.
\end{credits}

\bibliographystyle{splncs04}
\bibliography{MICCAI24_paper_w_supp}
%
% \newpage




% LNCS Homepage, \url{http://www.springer.com/lncs}. Last accessed 4
% \end{thebibliography}

\section*{Supplementary}
\setcounter{table}{0}
\begin{table}[ht!]
\centering
\begin{tabular}{|c|c|c|c|c|}
\hline
\rowcolor{light-gray}
 \textbf{Object} & \textbf{Feature} & \textbf{Dim} & \textbf{Equation} & \textbf{Description} \\
\hline
\hline
\rowcolor{light-blue}
\multicolumn{5}{|c|}{\textbf{Morphology Features (8)}} \\
\hline

\multirow{1}{*}{Nuc/Cell} & \multirow{1}{*}{Size} & \multirow{1}{*}{2} &  \multirow{1}{*}{$A=|O|$} & Area of object $O$. \\
 \hline
\multirow{2}{*}{Nuc/Cell} & \multirow{2}{*}{Roundness} & \multirow{2}{*}{2} & \multirow{2}{*}{$R=4\pi A/ C^2_p$} & Closeness to a circle.\\ 
 & & & & $C_p$, convex perimeter. \\
 %&  &  & & the shape of a segmentation \\
 %&  &  & & approaches that of a circle. \\
\hline
\multirow{1}{*}{Nuc/Cell} & \multirow{1}{*}{Elongation} & \multirow{1}{*}{2} & \multirow{1}{*}{$E=4A/\pi l_m^2$} & Area to major axis $l_m$ \\

\hline
\multirow{2}{*}{Nuc/Cell} & $1^{\text{st}}$ Hu & \multirow{2}{*}{2} & \multirow{2}{*}{$M_1 = \mu_{2,0} + \mu_{0,2}$} & Shape descriptor using  \\
 & Moment & & & moments, $\mu_{0,2}, \mu_{2,0}$ \\
 \hline
\hline
\rowcolor{light-blue}
\multicolumn{5}{|c|}{\textbf{Intensity Features (44)}} \\
%\hline
%Object & Feature & Dim & Equation & Description \\
\hline
\multirow{2}{*}{Nuc/Cell} & \multirow{2}{*}{MFI} & \multirow{2}{*}{6} & \multirow{2}{*}{$\text{MFI}_{ch}= \frac{1}{A} \sum_{p\in O} I_{ch}(p)$} & Mean Fluorescence Intensity\\ 
& & & & of object $O$ for channel $I_{ch}$\\
\hline
\multirow{2}{*}{Nuc/Cell} & Quartile & \multirow{2}{*}{24} & $\text{LQI}_{ch}$, $\text{MQI}_{ch}$, $\text{UQI}_{ch}$,  & Lower, median, upper, inter-\\
 & Intensities   &  & $\text{IQR}_{ch}\!=\!\text{UQI}_{ch}\!-\!\text{LQI}_{ch}$ & quartile range per channel \\
\hline
\multirow{2}{*}{Nuc/Cell} & \multirow{2}{*}{CK+ ratio} & \multirow{2}{*}{2} & \multirow{2}{*}{$CK_+\!=\!|I_{ck+}(p)_{p\in O}|/A$} & Fraction of pixels greater \\
 &  &  & & than CK cutoff in object\\
%\hline
 %Object & Feature & Dim & Equation & Description\\
\hline
\multirow{2}{*}{Nuc/Cell} & Channel & \multirow{2}{*}{6} & \multirow{2}{*}{Pearson's Correlation} & Similarity of pixels between \\
& Correlation & & & two channels within object\\
\hline
\multirow{2}{*}{Nuc/Cell} & Channel & \multirow{2}{*}{4} & Ranked-Weighted & Co-occurrence and corr. \\
& Co-localization &  & Co-localization & of two channels in object \\
\hline
CK & Statistics & 2 & $\mu_{ck}, \sigma_{ck}$ & Mean \& Std. Dev. of CK \\
\hline
\hline
\rowcolor{light-blue}
\multicolumn{5}{|c|}{\textbf{Texture Features (70)}} \\
\hline
\multirow{2}{*}{CK} & 2D Gabor & \multirow{2}{*}{32} & \multirow{2}{*}{$g(\lambda, \theta, \sigma, \psi, \gamma)$} & Frequency patterns in \\
& Filters & & & various orientations\\
\hline
\multirow{2}{*}{CK} & 2D Laws & \multirow{2}{*}{32} & Pairs of Level, Edge, & Spatial patterns in \\
& Filters & & Spot, and Ripple filters & various orientations\\
\hline
DAPI, & LBP & \multirow{3}{*}{6} & $LBP(p)\!=\!\sum_{i} s(n\textsubscript{i}-p)2^i$  & Local geometric pattern\\
CK, & Correlation & & $s(x)\!=\!1 \text{ if } x\!>\!1, \text{ else } 0$& encodings and similarity \\
CD45/31 & \& Mutal Info & & pixel $n_i$ around $p$ & between channel LBPs \\
%Images & \& Mutal Info & & around center, $p$ & LBP image pairs\\
\hline
% \multirow{4}{*}{CK Image} & \multirow{4}{*}{Blur Score} & \multirow{4}{*}{2} & Laplacian of Gaussian (LoG) -\\
% & & & measure if image is blurred \\
% & & & Difference of Gaussian (DoG) - \\
% & & & measure if image is blurred  \\
% \hline
% \multirow{2}{*}{CK Image} & Histogram of & \multirow{2}{*}{1} & HOG grid descriptor measures \\
% & Gradients & & the distribution of intensity gradients \\
% \hline
\end{tabular}
\caption{Summary of 122 Morphology, Intensity and Texture features extracted from 3-channel IF images (DAPI, CK, CD45/31) within objects of nucleus (Nuc.), cell or full image thumbnail.}
\label{table:features}
\end{table}

\end{document}